\documentclass[10pt,twocolumn,letterpaper]{article}

\usepackage{cvpr}
\usepackage{times}
\usepackage{epsfig}
\usepackage{graphicx}
\usepackage{amsmath}
\usepackage{amssymb}

\usepackage{caption}
\usepackage{subcaption}

\usepackage[ruled,vlined]{algorithm2e}
\usepackage{authblk}

\usepackage[pagebackref=true,breaklinks=true,letterpaper=true,colorlinks,bookmarks=false]{hyperref}

\cvprfinalcopy 

\def\cvprPaperID{****} 

\begin{document}

\title{3D Multi-Object Tracking for Autonomous Driving using a Kalman Filter}
\title{3D Multi-Object Tracking for Autonomous Driving}
\title{Good old Kalman Filter for 3D Multi-Object Tracking for Autonomous Driving}
\title{Probabilistic 3D Multi-Object Tracking for Autonomous Driving}

\author{Hsu-kuang Chiu$^1$, Antonio Prioletti$^2$, Jie Li$^2$, and Jeannette Bohg$^1$\\
$^1$Stanford University, $^2$Toyota Research Institute\\


}

\maketitle

\begin{abstract}
3D multi-object tracking is a key module in autonomous driving applications that provides a reliable dynamic representation of the world to the planning module.
In this paper, we present our on-line tracking method, which made the first place in the NuScenes Tracking Challenge, held at the AI Driving Olympics Workshop at NeurIPS 2019. Our method estimates the object states by adopting a Kalman Filter. We initialize the state covariance as well as the process and observation noise covariance with statistics from the training set. We also use the stochastic information from the Kalman Filter in the data association step by measuring the Mahalanobis distance between the predicted object states and current object detections.
Our experimental results on the NuScenes validation and test set show that our method outperforms the AB3DMOT baseline method by a large margin in the Average Multi-Object Tracking Accuracy (AMOTA) metric. Our code will be available soon.\footnotemark 
\let\thefootnote\relax\footnotetext{\url{https://github.com/eddyhkchiu/mahalanobis_3d_multi_object_tracking}. }
\end{abstract}

\section{Introduction}
3D multi-object tracking is essential for autonomous driving. Its aim is to  estimate the location, orientation, and scale of all the objects in the environment over time. By taking temporal information into account, a tracking module can  filter outliers in frame-by-frame object detectors and be robust to partial or full occlusions. Thereby, it promises to identify the trajectories of different categories of moving objects, such as pedestrians, bicycles, and cars. The resulting trajectories may then be used to infer motion patterns and driving behaviours for improved forecasting. This in turn helps planning to enable autonomous driving.

In this paper, we approach the 3D multi-object tracking problem with a Kalman Filter~\cite{kalman1960filter}. We model the state of each object with its 3D position, orientation and scale as well as linear and angular velocity. For the prediction step, we use a process model with constant linear and angular velocity. We model the unknown accelerations as Gaussian random variables. For the update step, we consider the detections provided by an object detector as measurements. Similar to the process model, we also model measurement noise as Gaussian random variables.
To ensure robustness in multi-object tracking we found the following two steps to be essential: (i) we employ the Mahalanobis distance \cite{mahalanobis1936distance} for outlier detection and data association between predicted and actual object detections; (ii) we estimate the covariance matrices of the initial state and of the process and observation noise from the training data. 

For data association between the predicted and actual object detections, we found that using the Mahalanobis distance \cite{mahalanobis1936distance} is better than using the {\em 3D Intersection-Over-Union\/} (3D-IOU) as in the AB3DMOT \cite{weng2019ab3dmot} baseline and other previous works. 
Differently from the 3D-IOU, the Mahalanobis distance takes into account the uncertainty about the predicted object state as provided by the Kalman Filter in form of the state covariance matrix. Moreover, the Mahalanobis distance can provide distance measurement even when prediction and detection do not overlap. In this case, the 3D-IOU gives zero which prevents any data association. However, non-overlapping detections and predictions are highly common in driving scenarios due to sudden accelerations and for smaller objects such as pedestrians and bicycles.

Correctly choosing the initial state and noise covariance matrices is fundamental for filter convergence. Moreover, the reliability of Mahalanobis distance directly depends on the choice of these values which thereby influence the quality of data association. We extract the statistics of the training data to perform this initialization. This also ensures that our experiments on the validation and test set do not use any future or ground-truth information.

We evaluate our approach in the NuScenes Tracking Challenge \cite{caesar2019nuscenes} using the provided MEGVII \cite{zhu2019megvii} detection results as measurements. Our proposed method outperforms the AB3DMOT \cite{weng2019ab3dmot} baseline by a large margin in terms of the {\em Average Multi-Object Tracking Accuracy\/} (AMOTA) and made the first place in the NuScenes Tracking Challenge.

\section{Related Work}

\subsection{3D Object Detection}
The 3D object detection component provides object bounding boxes for each frame as measurements to  3D multi-object tracking systems. Therefore, the quality of the 3D object detector is essential for the final tracking accuracy. In general, most Lidar based 3D object detection methods belong to one of two categories: voxel- or point-based methods. Voxel-based methods first divide the 3D space into equally-sized 3D voxels to generate 3D feature tensors based on the points inside each voxel. Then the feature tensors are fed to 3D CNNs to predict the object bounding boxes. Point-based methods do not need the quantization step. Those methods directly apply PointNet++ \cite{qi2017pointnetplusplus} on the raw point cloud data for detecting the objects in the 3D space. A more recent work that achieves state-of-the-art in the NuScenes Detection Challenge \cite{caesar2019nuscenes} is the MEGVII \cite{zhu2019megvii} model. This model is a voxel-based method and utilizes sparse 3D convolution to extract the semantic features. Then a region proposal network and class-balanced multi-head networks are used for final object detection. We use the MEGVII \cite{zhu2019megvii} detection results as measurements  for the 3D multi-object tracker.

\subsection{3D Multi-Object Tracking} \label{sub_sec:3d_multi_object_tracking}
Several 3D multi-object tracking methods are extensions of 2D tracking methods. Weng et al.~\cite{weng2019ab3dmot} propose AB3DMOT, a simple yet effective on-line tracking method based on a 3D Kalman Filter. Hu et al. \cite{hu2019joint} combine LSTM-based 3D motion estimation with 2D image deep feature association for solving the 3D tracking problem. Different from the above methods that only rely on the image or point cloud sensor input data, Argoverse \cite{argoverse2019} further uses the map information, such as lanes and drivable areas, to improve 3D multi-object tracking accuracy in driving scenarios. 

In our work, we use AB3DMOT \cite{weng2019ab3dmot} as our baseline. AB3DMOT uses a 3D Kalman Filter for tracking. Each of their Kalman Filter states includes the center position, rotation angle, length, width, height of the object bounding box, and center velocity, while excluding the angular velocity. Their Kalman Filter covariance matrices are identity matrices multiplied with a heuristically chosen scalar. Moreover, AB3DMOT uses the 3D-IOU as the affinity function and the Hungarian algorithm for data association.

In our approach, we propose to utilize the Mahalanobis distance~\cite{mahalanobis1936distance} for measuring affinity between predictions and detections with or without direct overlapping. This distance takes the uncertainty in the predictions into account and is standard practice for outlier detection in filtering method~\cite{Thrun:2005:PR:1121596}. In our approach, we also estimate the state and noise covariance matrices from the statistics of the training data. In experiments, we quantitatively show that these two measures improve the performance of the multi-object tracker by a large margin. We also included angular velocity in the state and found that qualitatively the trajectories look more accurate especially in terms of object orientation. 

\section{A Kalman Filter for Multi-Object Tracking}
\begin{figure*}[t!]
    \centering
    \includegraphics[width=1.0\linewidth]{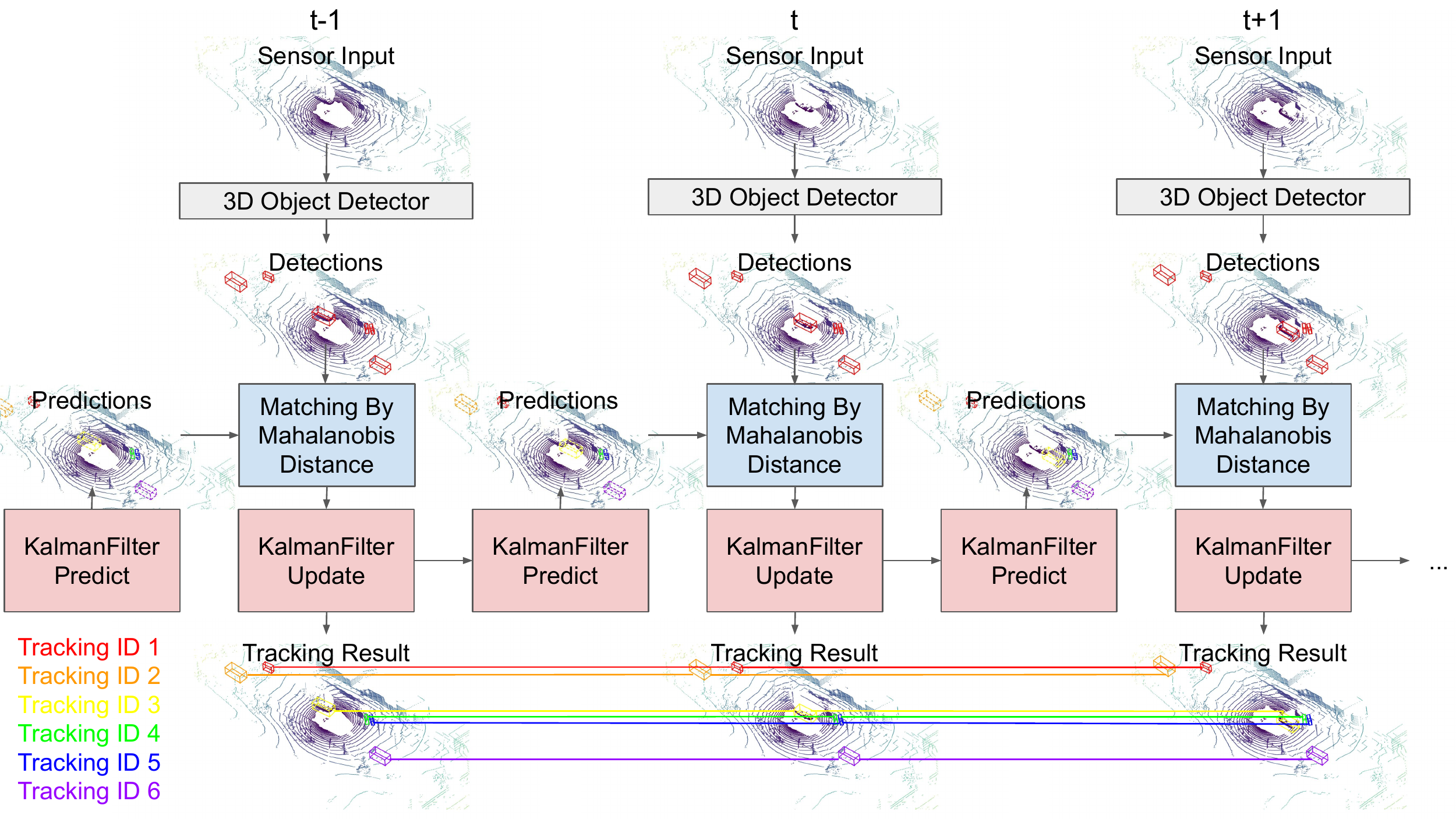}
    \caption{Architecture Overview. We use 3D object detection results as measurements. At each timestep, we use the Mahalanobis distance \cite{mahalanobis1936distance} to compute the distance between object detections and predictions. Given this distance, we perform data association. The Kalman Filter \cite{kalman1960filter} then updates the current state estimates. It uses a constant velocity model for predicting the mean and covariance of the state in the next time step.
    }
    \label{fig:arch}
\end{figure*}

In this section, we introduce our proposed 3D multi-object tracking algorithm built upon a Kalman Filter \cite{kalman1960filter}.
In the prediction step, we use a process model assuming constant linear and angular velocity. We assume that an object detector provides frame-by-frame measurements to the filter. These detections are matched to the predicted detections to then update the current object state estimates.
The overall architecture is shown in Fig.~\ref{fig:arch}. In the following sections, we first model the dynamical system we are estimating and then describe how we tune the open parameters in the filter as well as how we perform data association.

\subsection{Object State}\label{sec:object_state}
We model each object's state with a tuple of 11 variables:
\begin{equation}
{\bf s}_t = (x, y, z, a, l, w, h, d_x, d_y, d_z, d_a)^T,
\end{equation}
where $(x, y, z)$ represent the 3D object center position, $a$ represents the object orientation about the z-axis, $(l, w, h)$ represent the length, width, and height of the object's bounding box, and $(d_x, d_y, d_z, d_a)$ represent the change of $(x, y, z, a)$ from the previous frame to the current frame. The last four variables are the linear and angular velocity of the object center multiplied by a constant $\Delta t$. Please note, that we are tracking multiple objects and therefore maintain $M$ such states, one per tracked object in the scene.

\subsection{Process Model}\label{sec:process_model}
We model the dynamics of the moving objects using the following process model: 

\begin{tabular}{ l l }
\(\displaystyle \hat{x}_{t+1} = x_t + d_{x_t} + q_{x_t}, \) & \(\displaystyle \hat{d}_{x_{t+1}} = d_{x_t} + q_{d_{x_t}} \)  \\
\(\displaystyle \hat{y}_{t+1} = y_t + d_{y_t} + q_{y_t}, \) & \(\displaystyle \hat{d}_{y_{t+1}} = d_{y_t} + q_{d_{y_t}} \)  \\
\(\displaystyle \hat{z}_{t+1} = z_t + d_{z_t} + q_{z_t}, \) & \(\displaystyle \hat{d}_{z_{t+1}} = d_{z_t} + q_{d_{z_t}} \)  \\
\(\displaystyle \hat{a}_{t+1} = a_t + d_{a_t} + q_{a_t}, \) & \(\displaystyle \hat{d}_{a_{t+1}} = d_{a_t} + q_{d_{a_t}} \)  \\
\(\displaystyle \hat{l}_{t+1} = l_t \) \\
\(\displaystyle \hat{w}_{t+1} = w_t \) \\
\(\displaystyle \hat{h}_{t+1} = h_t \) \\
\end{tabular}

where we model the unknown linear and angular acceleration as random variables $(q_{x_t}, q_{y_t}, q_{z_t}, q_{a_t})$ and $(q_{d_{x_t}}, q_{d_{y_t}}, q_{d_{z_t}}, q_{d_{a_t}})$ that follow a Gaussian distribution with zero mean and covariance $\mathbf{Q}$. We assume constant linear and angular velocity as well as constant object dimensions, i.e. they do not change during the prediction step. Note that those variable may change during the update step.

We can then write the Kalman Filter prediction step in matrix form as follows:
\begin{align}
\label{eq:predicted_mean}
    {\bf \hat{\mu}}_{t+1} &= {\bf A} \mathbf{\mu}_t\\
    \label{eq:predicted_covariance}
    \hat{\Sigma}_{t+1} & = {\bf A} \Sigma_t {\bf A}^T + {\bf Q}
\end{align}
where ${\bf \mu}_t$ is the estimated mean of the true state $s$ at time $t$, and $\hat{\mu}_{t+1}$ is the predicted state mean at time $t+1$. The matrix ${\bf A}$ is the state transition matrix of the process model. The matrix $\Sigma_t$ is the state covariance at time $t$, and $\hat{\Sigma}_{t+1}$ is the predicted state covariance at time $t+1$. 

\subsection{Observation Model}
We assume that an object detector provides us with $N$ frame-by-frame measurements ${\bf o}$ of object states, i.e. position, orientation and bounding box scale. The number of detections may differ from the number of tracked, individual objects. For now, let us assume that we already matched one of the detections to an object state. In the next section, we provide detail on data association. 

As detections are direct measurements of parts of the state ${\bf \mu}$, the linear observation model has the following matrix form: ${\bf H}_{7 \times 11} = [{\bf I}~{\bf 0} ]$. Similar to the process model, we assume observation noise follows a Gaussian distribution with zero mean and covariance $\mathbf{R}$. Using this observation model and the predicted object state ${\bf \hat{\mu}}_{t+1}$, we can predict the next measurement ${\bf \hat{o}}_{t+1}$ and innovation covariance ${\bf S}_{t+1}$ that represents the uncertainty of the predicted object detection:
\begin{align}
    {\bf \hat{o}}_{t+1} & = {\bf H} {\bf \hat{\mu}}_{t+1}\\
    {\bf S}_{t+1} & = {\bf H} \hat{\Sigma}_{t+1} {\bf H}^T + {\bf R}
\end{align}

We will discuss how we estimate the value of $\Sigma_0$, $\mathbf{Q}$, and $\mathbf{R}$ in Section \ref{sub_sec:covariance_estimation}.

\subsection{Data Association}
We are using an object detector to provide the Kalman Filter with $N$ measurements. As the detector results can be noisy, we need to design a data association mechanism to decide which detection to pair with a predicted object state and which detections to treat as outliers. Previous work~\cite{weng2019ab3dmot} has used 3D-IOU to measure the affinity between predictions and detections. We adopt the fairly standard practice~\cite{Thrun:2005:PR:1121596} of using the Mahalanobis distance \cite{mahalanobis1936distance} instead. This distance $m$ measures the difference between predicted detections ${\bf H}\hat{\mu}_{t+1}$ and actual detections $\mathbf{o}_{t+1}$ weighted by the uncertainty about the prediction as expressed through the innovation covariance ${\bf S}_{t+1}$:
\begin{equation}
m = \sqrt{({\bf o}_{t+1} - {\bf H}\hat{\mu}_{t+1})^{T} {{\bf S}_{t+1}}^{-1} ({\bf o}_{t+1} - {\bf H}\hat{\mu}_{t+1})}.
\end{equation}

We also adopt the orientation correction approach from the AB3DMOT \cite{weng2019ab3dmot} baseline. Specifically, when the angle difference between the detection and prediction is between 90 and 270 degrees, we rotate the prediction's angle by 180 degrees before calculating the Mahalanobis distance. Large angle difference like that usually stems from the detector that outputs an incorrect facing direction of the object. Furthermore, it is unlikely that the object makes such a large turn in the short time duration between consecutive frames. In our experiments, we show that the Mahalanobis distance provides better tracking performance than the 3D-IOU.

Given the distances between all predictions and detections, we solve a bipartite matching problem to find the optimal pairing.
Specifically, we employ a greedy algorithm with an upper bound threshold value to solve this problem. The algorithm is described in detail in  Algorithm~\ref{algo:greedy}. Compared to the Hungarian algorithm as used in the AB3DMOT baseline~\cite{weng2019ab3dmot}, the greedy approach performs better as shown in our ablative analysis in Section \ref{sub_sec:quantitative_results_and_ablations}.

\begin{algorithm}[h]
\SetAlgoLined
\KwIn{\\
$M$ predicted means and innovation covariance matrices, one per tracked object: $P = \{(\hat{\mu}^{[1]}, S^{[1]}), (\hat{\mu}^{[2]}, S^{[2]}), \ldots, (\hat{\mu}^{[M]}, S^{[M]})\}$. \\
$N$ detections: $D = \{o^{[1]}, o^{[2]}, \ldots, o^{[N]}\}$. \\
A threshold $T$ as the upper bound of matched pair's Mahalanobis distance.\\
}
\KwOut{\\
List of bipartite matched pair indices sorted by the Mahalanobis distance.
}

\SetKwInput{KwInit}{Initialization}
\KwInit{\\
$List \gets \emptyset$ \\
$MatchedP \gets \emptyset$ \\
$MatchedD \gets \emptyset$ \\
$Distance \gets array[M][N]$}
\For{$i \gets 1$ \textbf{to} $M$}{
  \For{$j \gets 1$ \textbf{to} $N$}{
    $Distance[i][j] \gets MahalanobisDistance((\hat{\mu}^{[i]}, S^{[i]}), o^{[j]})$
  }
}
$Pairs \gets IndexPairsSortByValue(Distance)$ \\
\For{$k \gets 1$ \textbf{to} $length(Pairs)$}{
  $(m, n) \gets Pairs[k]$ \\
  \If{$m \not \in MatchedP$ \textbf{and} $n \not \in MatchedD$} {
    \uIf{$Distance[m][n] < T$} {
      $List \gets append(List, (m, n))$ \\
      $MatchedP \gets MatchedP \cup \{m\}$ \\
      $MatchedD \gets MatchedD \cup \{n\}$ \\
    } 
    \Else {
      \textbf{break} \\
    }
  }
}
\Return{$List$}
\caption{Greedy Algorithm for Data Association at time $t$}
\label{algo:greedy}
\end{algorithm}

\subsection{Kalman Filter Update Step}
Given the matched pairs of detections and predictions, we can now update the predicted state mean and covariance at time $t+1$ by using the following equations:
\begin{align*}
{\bf K}_{t+1} & = \hat{\Sigma}_{t+1} {\bf H}^T {\bf S}_{t+1}^{-1}\\
    \mu_{t+1} & = \hat{\mu}_{t+1} + {\bf K}_{t+1} ({\bf o}_{t+1} - {\bf H} \hat{\mu}_{t+1}) \\
    \Sigma_{t+1} & = ({\bf I} - {\bf K}_{t+1}{\bf H}) \hat{\Sigma}_{t+1}  
\end{align*}
where ${\bf K}$ refers to the Kalman Gain
and the matrix ${\bf I}$ is an identity matrix. 


We also follow the aforementioned orientation correction in the update step. We adopt the birth-and-death memory module from the AB3DMOT \cite{weng2019ab3dmot} baseline: we initialize a track after having matches for 3 consecutive frames. And we terminate a track when it does not match any detection for 2 consecutive frames.

\subsection{Covariance Matrices Estimation} \label{sub_sec:covariance_estimation}
Rather than using the identity matrices and heuristically chosen scalars  to build the covariance matrices of the Kalman Filter as in AB3DMOT \cite{weng2019ab3dmot}, we use the statistics of the training set data to estimate the initial state covariance, the process and observation noise covariance. Note that we did not use the statistics from the validation or test set, to make sure that our experiment does not use any future or ground-truth information in the evaluation. 

Specifically, our process noise models the unknown linear and angular accelerations. Therefore, we analyse the variance in the ground truth accelerations in the training data set. Let us denote the training set's ground-truth object center positions and rotation angles as $(x_t^{[m]}, y_t^{[m]}, z_t^{[m]}, a_t^{[m]})$ for timestamp $t \in \{1\cdots T\}$ and object index $m \in \{1 \cdots M\}$. We model the process noise covariance as a diagonal matrix where each element is associated to the center positions and rotation angles $(Q_{xx}, Q_{yy}, Q_{zz}, Q_{aa})$ and estimated as follows:
\begin{align}\label{eq:Q_xx}
    Q_{xx} & = Var( (x_{t+1}^{[m]} - x_t^{[m]}) - (x_t^{[m]} - x_{t-1}^{[m]}) )\\
    Q_{yy} & = Var( (y_{t+1}^{[m]} - y_t^{[m]}) - (y_t^{[m]} - y_{t-1}^{[m]}) )\\
    Q_{zz} & = Var( (z_{t+1}^{[m]} - z_t^{[m]}) - (z_t^{[m]} - z_{t-1}^{[m]}) )\\
    Q_{aa} & = Var( (a_{t+1}^{[m]} - a_t^{[m]}) - (a_t^{[m]} - a_{t-1}^{[m]}) )
\end{align}
The above variances are calculated over $m \in \{1, ..., M\}$ and $t \in \{2, ..., T-1\}$. 
The $Q$'s elements associate to the center velocity and angular velocity $(Q_{d_xd_x}, Q_{d_yd_y}, Q_{d_zd_z}, Q_{d_ad_a})$ are estimated in the same way as follows:
\begin{equation}\label{eq:Q_dxdx}
(Q_{d_xd_x}, Q_{d_yd_y}, Q_{d_zd_z}, Q_{d_ad_a}) = (Q_{xx}, Q_{yy}, Q_{zz}, Q_{aa})
\end{equation}

One might think that the above estimation seems to double count the acceleration. However, the above estimation is actually reasonable based on our process model definition. For example, consider the $x$ component of the state and its velocity-related component $d_x$ in the process model defined in Section \ref{sec:process_model}:
\begin{align}
    \label{eq:predicted_position}
    & \hat{x}_{t+1} = x_t + d_{x_t} + q_{x_t} \\
    \label{eq:predicted_velocity}
    & \hat{d}_{x_{t+1}} = d_{x_t} + q_{d_{x_t}}
\end{align}

To estimate the two noise terms $q_{x_t}$ and $q_{d_{x_t}}$, we have:
\begin{align}
     & q_{x_t} = \hat{x}_{t+1} - x_t - d_{x_t} \\
     & q_{d_{x_t}} = \hat{d}_{x_{t+1}} - d_{x_t}
\end{align}
where the predicted state components $\hat{x}_{t+1}$ and $\hat{d}_{x_{t+1}}$ can be estimated using the ground-truth state components $x_{t+1}$ and $d_{x_{t+1}}$. The velocity-related components $d_{x_{t+1}}$ and $d_{x_t}$ can be approximated as $x_{t+1} - x_t$ and $x_t - x_{t-1}$ based on our state definition in Section \ref{sec:object_state}. And we can derive the equations as follows:
\begin{align}
     q_{x_t} 
     & \approx {x}_{t+1} - x_{t} - d_{x_t} \\
     & \approx (x_{t+1} - x_{t}) - (x_{t} - x_{t-1}) \\
     q_{d_{x_t}} 
     & \approx {d}_{x_{t+1}} - d_{x_t} \\
     & \approx (x_{t+1} - x_{t}) - (x_{t} - x_{t-1})
\end{align}

The above approximation explains why we use the variance of accelerations to estimate the process model noise covariance from equation \ref{eq:Q_xx} to \ref{eq:Q_dxdx}. 

Additionally, including the acceleration noise in both prediction equations in \ref{eq:predicted_position} and \ref{eq:predicted_velocity} also adds robustness to the data association. On the contrary, only including the acceleration to the velocity prediction in equation \ref{eq:predicted_velocity} will underestimate the uncertainty when predicting the next position. Consider the case that there is a very large real acceleration in the current time step which could not be accounted for in the previously estimated velocity. In this case, we will have large uncertainty in predicting the next velocity. But we will only have small uncertainty in predicting the next position if we do not include the acceleration noise to the position prediction equation. By adding this additional acceleration noise in position prediction, we increase the predicted uncertainty of position. And that is used within the Mahalanobis distance and therefore the data association becomes more generous for matching and more robust. Similar reasoning also applies to other state variables.

And for the elements related to the length, width, height, and other non-diagonal elements in $Q$, we assume their variances to have value $0$.

Our observation noise models the error in the object detector. Therefore, we analyse the error variance between ground-truth object poses and detections in the training set to then choose the diagonals entries of ${\bf R}$ and the initial state covariance $\Sigma_0$.
For this, we first find the matching pairs of the detection bounding boxes and the ground-truth by using the matching criteria that the 2D center distance is less than 2 meters. Given the matched pairs of the detections and the ground-truth $(D_t^{[k]}, G_t^{[k]})$ for timestamp $t \in \{ 1\cdots T\}$ and matched pair index $k \in \{ 1 \cdots K\}$, where 
\begin{align}
& D_t^{[k]} = (D_{x_t}^{[k]}, D_{y_t}^{[k]}, D_{z_t}^{[k]}, D_{a_t}^{[k]}, D_{l_t}^{[k]}, D_{w_t}^{[k]}, D_{h_t}^{[k]}) \\
& G_t^{[k]} = (G_{x_t}^{[k]}, G_{y_t}^{[k]}, G_{z_t}^{[k]}, G_{a_t}^{[k]}, G_{l_t}^{[k]}, G_{w_t}^{[k]}, G_{h_t}^{[k]}) 
\end{align}
we estimate the elements of the observation noise covariance matrix $R$ as follows:
\begin{align}
& R_{xx} = Var( D_{x_t}^{[k]} - G_{x_t}^{[k]}) \\    
& R_{yy} = Var( D_{y_t}^{[k]} - G_{y_t}^{[k]}) \\    
& R_{zz} = Var( D_{z_t}^{[k]} - G_{z_t}^{[k]}) \\    
& R_{aa} = Var( D_{a_t}^{[k]} - G_{a_t}^{[k]}) \\    
& R_{ll} = Var( D_{l_t}^{[k]} - G_{l_t}^{[k]}) \\    
& R_{ww} = Var( D_{w_t}^{[k]} - G_{w_t}^{[k]}) \\    
& R_{hh} = Var( D_{h_t}^{[k]} - G_{h_t}^{[k]})    
\end{align}
The non-diagonal entries of ${\bf R}$ are all zero. We set $\Sigma_0 = {\bf R}$ as we initialize the multi-object tracker with the initial detection results.

\section{Experiment Results}
\begin{table*}[t!]
\small
\caption{Tracking results for the validation set of NuScenes \cite{caesar2019nuscenes}: evaluation in terms of overall AMOTA and individual AMOTA for each object category in comparison with the AB3DMOT \cite{weng2019ab3dmot} baseline method, and variations of our method. In each column, the best obtained results are typeset in boldface. (*Our baseline implementation of applying AB3DMOT \cite{weng2019ab3dmot} on the MEGVII \cite{zhu2019megvii} detection result.)\vspace{-15pt}}
\label{tab:val_results}
\begin{center}
\begin{tabular}{ l|cccccccc}
  \hline
  Method & Overall & bicycle & bus & car & motorcycle & pedestrian & trailer & truck \\
  \hline
  AB3DMOT \cite{weng2019ab3dmot} & 17.9 & 0.9 & 48.9 & 36.0 & 5.1 & 9.1 & 11.1 & 14.2 \\
  AB3DMOT \cite{weng2019ab3dmot} * & 50.9 & 21.8 & 74.3 & 69.4 & 39.0 & 58.7 & 35.3 & 58.1 \\
  \hline
  Ours w/ 3D-IOU, threshold 0.01 & 52.7 & 23.2 & 73.9 & 72.1 & 40.4 & 66.7 & 34.4 & 58.3 \\
  Ours w/ 3D-IOU, threshold 0.1 & 49.2 & 22.3 & \textbf{74.4} & 68.2 & 38.9 & 47.1 & 35.9 & 57.8 \\
  Ours w/ 3D-IOU, threshold 0.25 & 43.9 & 21.3 & 73.9 & 63.3 & 35.1 & 21.6 & \textbf{36.9} & 54.9 \\
  \hline
  Ours w/ Hungarian algorithm & 49.8 & 24.2 & 68.4 & 63.9 & 42.9 & 70.0 & 27.6 & 52.0 \\
  Ours w/ default covariance & 41.7 & 11.2 & 57.0 & 56.8 & 37.8 & 63.7 & 23.4 & 41.7 \\
  Ours w/o angular velocity & \textbf{56.1} & \textbf{27.2} & 74.1 & \textbf{73.5} & \textbf{50.7} & \textbf{75.5} & 33.8 & \textbf{58.1} \\
  \hline
  Ours & \textbf{56.1} & \textbf{27.2} & 74.1 & \textbf{73.5} & 50.6 & \textbf{75.5} & 33.7 & 58.0\\
  \hline
\end{tabular}
\end{center}
\vspace{-20pt}
\end{table*}


\begin{table}[t!]
\small
\caption{Tracking results for the test set of NuScenes  \cite{caesar2019nuscenes}. The full tracking challenge leaderboard will be released to public soon by the organizer. \vspace{-15pt}}
\label{tab:test_results}
\begin{center}
\begin{tabular}{ c|l|c}
  \hline
  Rank & Team Name & AMOTA \\
  \hline
  \textbf{1} & StanfordIPRL-TRI (Ours) & \textbf{55.0} \\
  \hline
  2 & VV\_team & 37.1 \\
  \hline
  3 & CenterTrack & 10.8 \\
  \hline
  baseline & AB3DMOT \cite{weng2019ab3dmot} & 15.1 \\
  \hline
\end{tabular}
\end{center}
\vspace{-20pt}
\end{table}

\subsection{Evaluation Metrics}
We follow the NuScenes Tracking Challenge \cite{caesar2019nuscenes} and use the {\em Average Multi-Object Tracking Accuracy\/} (AMOTA) as the main evaluation metric. AMOTA is defined as follows:
{\small
\begin{equation}
    AMOTA = \frac{1}{n-1} \sum_{r \in \{ \frac{1}{n-1},  \frac{2}{n-1}, ... , 1\} } MOTAR,
\end{equation}
}
where $n$ is the number of evaluation sample points, and $r$ is the evaluation targeted recall. The MOTAR is the Recall-Normalized Multi-Object Tracking Accuracy, defined as the follows:
{\small
\begin{equation}
    MOTAR = max(0, 1 -  \frac{IDS_r + FP_r + FN_r - (1-r) * P}{r * P}),
\end{equation}
}
where $P$ is the number of ground-truth positives, $IDS_r$ is the number of identity switches, $FP_r$ is the number of false positives, and $FN_r$ is the number of false negatives.

\subsection{Baseline Evaluation}
We use AB3DMOT \cite{weng2019ab3dmot} as the baseline, as described earlier in Section \ref{sub_sec:3d_multi_object_tracking}. We report the AB3DMOT's tracking result for the NuScenes validation set in the first row of Table \ref{tab:val_results} as reported by the NuScenes Tracking Challenge \cite{caesar2019nuscenes}. Additionally, we adopted the AB3DMOT \cite{weng2019ab3dmot} open-source code on the MEGVII \cite{zhu2019megvii} detection results, and generate a better baseline tracking result, as reported in the second row of Table \ref{tab:val_results}. Currently, we do not know why the AMOTA numbers are different for the two implementations. 

\subsection{Quantitative Results and Ablations}\label{sub_sec:quantitative_results_and_ablations}
We report our method's results on the validation set in Table \ref{tab:val_results}.
We also include the AB3DMOT \cite{weng2019ab3dmot} baseline validation result in Table \ref{tab:val_results}. We can see that our method outperforms the official AB3DMOT baseline by a large margin (38.2\%) in terms of the overall AMOTA. Our method also achieves higher overall AMOTA compared with our baseline implementation of applying AB3DMOT \cite{weng2019ab3dmot} on the MEGVII \cite{zhu2019megvii} detection result by 5.2\%.

Additionally, we perform an ablation study by replacing different components of our method by the associate components of AB3DMOT \cite{weng2019ab3dmot}. We report the results in Table \ref{tab:val_results}. We can see that our proposed Mahalanobis distance-based data association method outperforms the 3D-IOU methods, especially in the categories of small objects, such as the bicycle, the motorcycle, and the pedestrian. For those objects, their 3D-IOU could be 0 even if the prediction and the detection are very close but do not overlap. In such cases, the 3D-IOU method will miss the match. However, our proposed Mahalanobis distance method can still correctly track the objects because this method still provides distance measurements even when the 3D-IOU is zero. The Mahalanobis distance also takes the uncertainty about the prediction into account as estimated by the Kalman Filter.

We also find that the greedy algorithm performs better than the Hungarian algorithm during the data association process. Our data-driven covariance matrix estimation outperforms the heuristic choices when using our Mahalanobis distance-based tracking method.

\begin{figure*}[h!]
        \centering
        \begin{subfigure}[b]{0.3\textwidth}
            \centering
            \includegraphics[width=\textwidth]{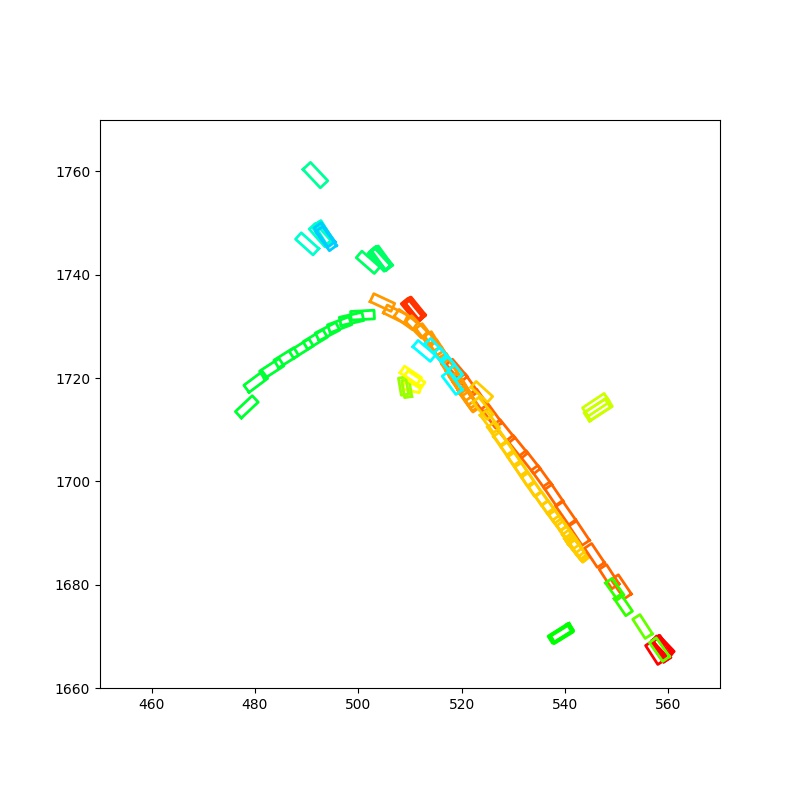}
            \caption[]%
            {{\small AB3DMOT \cite{weng2019ab3dmot}}}    
            \label{fig:car_weng2019ab3dmot_0}
        \end{subfigure}
        \hfill
        \begin{subfigure}[b]{0.3\textwidth}  
            \centering 
            \includegraphics[width=\textwidth]{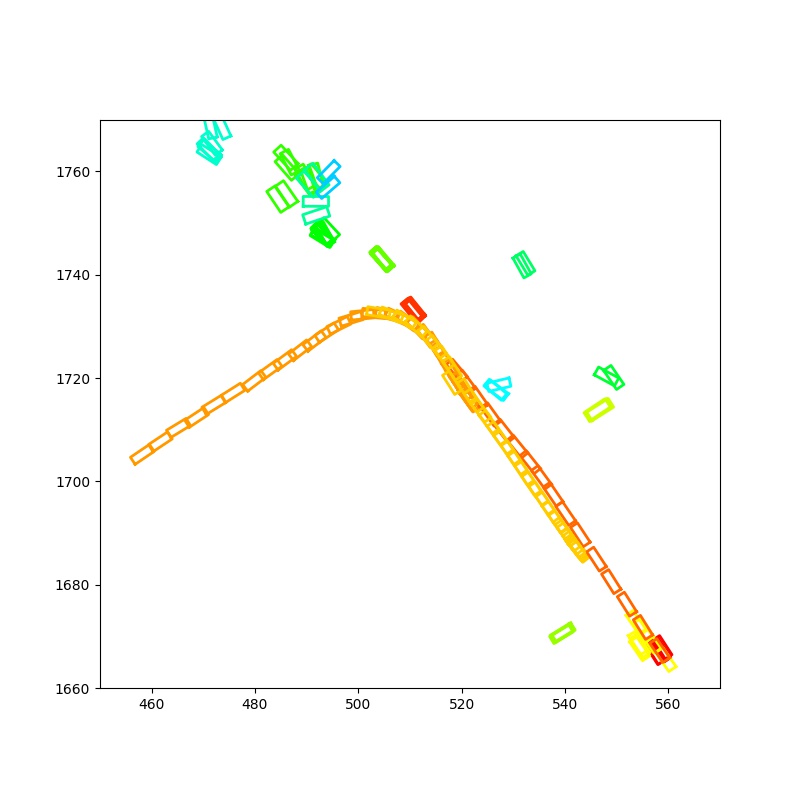}
            \caption[]%
            {{\small Ours}}    
            \label{fig:car_ours}
        \end{subfigure}
        \hfill
        \begin{subfigure}[b]{0.3\textwidth}
            \centering 
            \includegraphics[width=\textwidth]{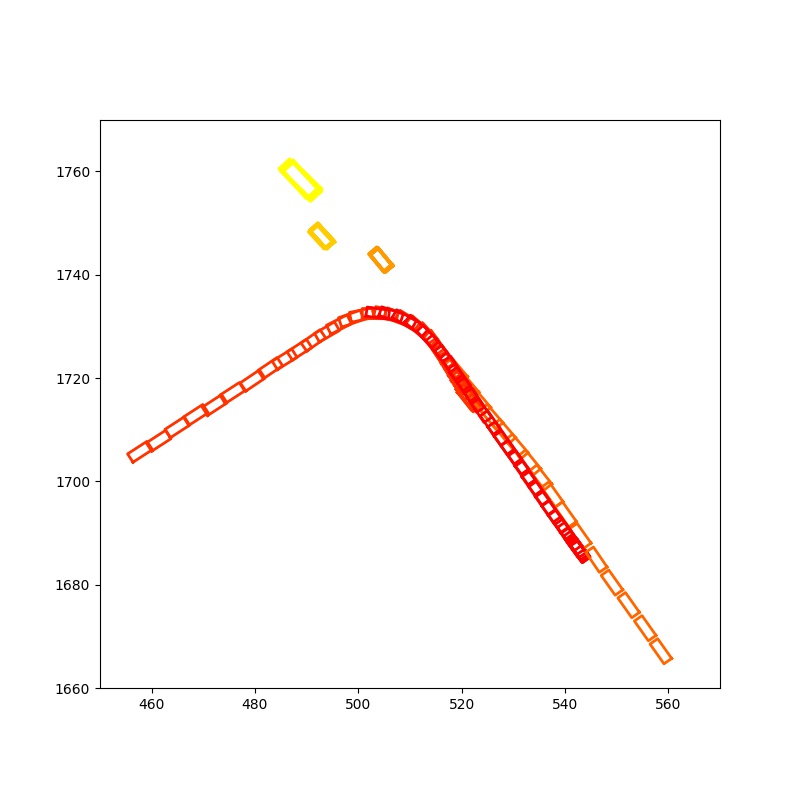}
            \caption[]%
            {{\small Ground-truth}}    
            \label{fig:car_gt}
        \end{subfigure}
        \vskip\baselineskip
        \begin{subfigure}[b]{0.3\textwidth}
            \centering 
            \includegraphics[width=\textwidth]{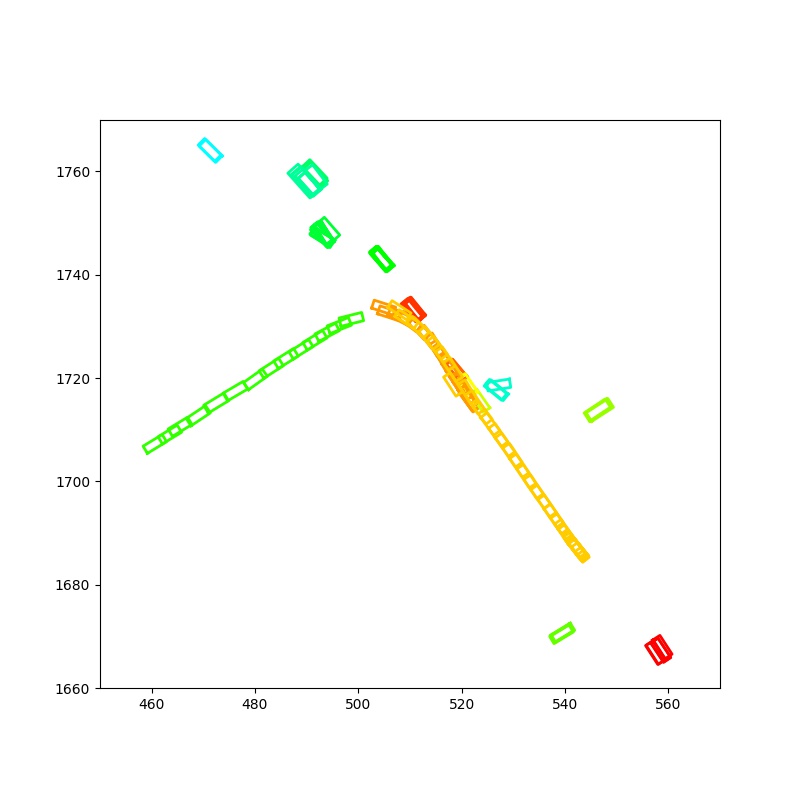}
            \caption[]%
            {{\small Our AB3DMOT \cite{weng2019ab3dmot} baseline}}    
            \label{fig:car_weng2019ab3dmot_1}
        \end{subfigure}
        \hfill
        \begin{subfigure}[b]{0.3\textwidth}
            \centering 
            \includegraphics[width=\textwidth]{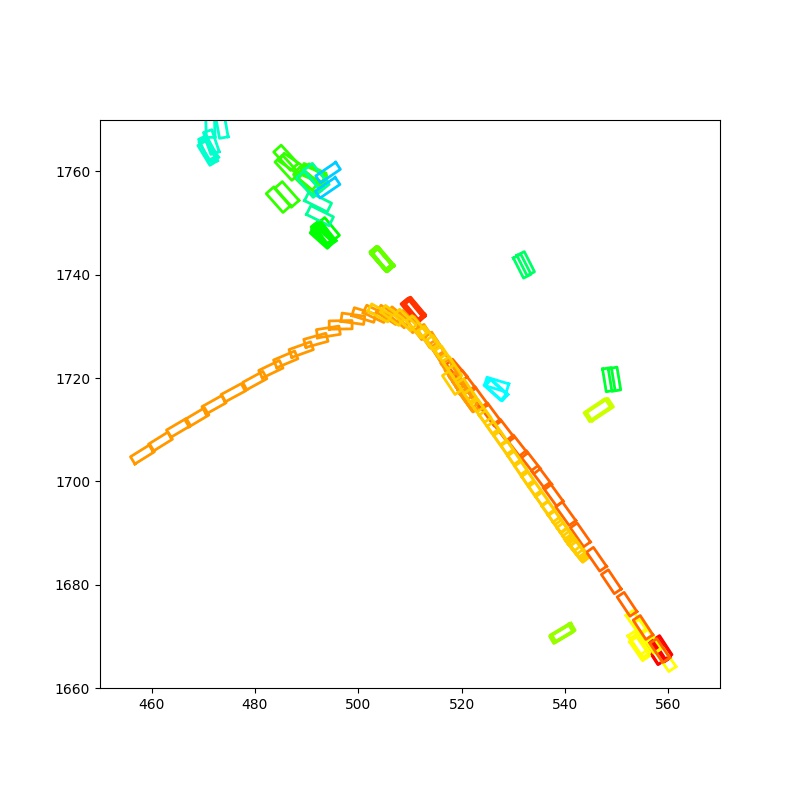}
            \caption[]%
            {{\small Ours without angular velocity}}    
            \label{fig:car_ours_without_angular_velocity}
        \end{subfigure}
        \hfill
        \begin{subfigure}[b]{0.3\textwidth}
            \centering 
            \includegraphics[width=\textwidth]{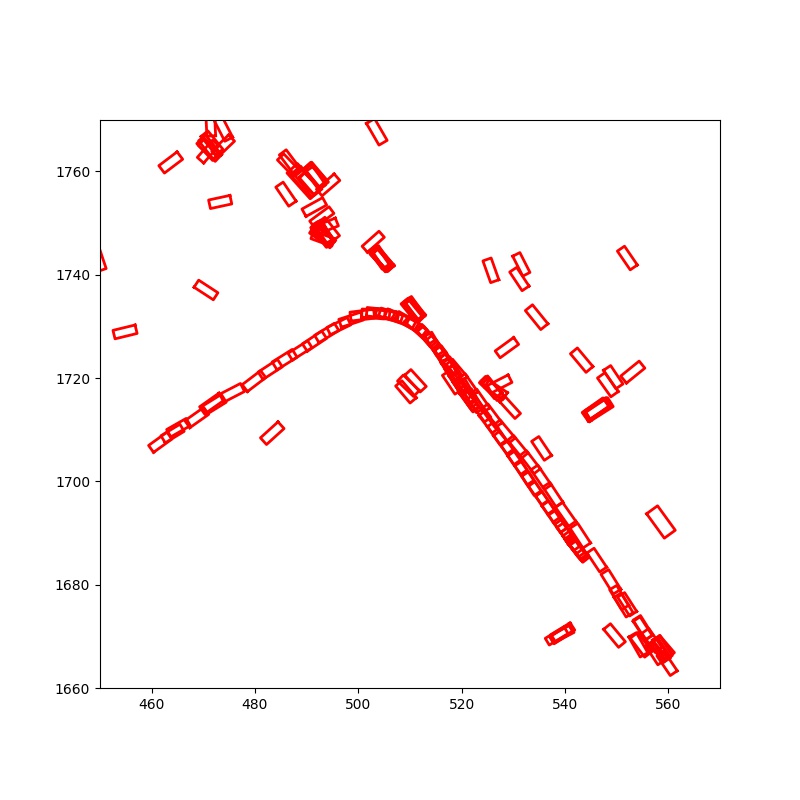}
            \caption[]%
            {{\small Input detection from MEGVII \cite{zhu2019megvii}}}    
            \label{fig:car_detection}
        \end{subfigure}
        \caption[]
        {\small Bird-eye-view tracking visualization of cars} 
        \label{fig:visual_car}
    \end{figure*}

One interesting finding of the ablation analysis is that excluding the angular velocity from the Kalman Filter state does not decrease the quantitative tracking performance in terms of the AMOTA. That is because the NuScenes tracking evaluation procedure uses 2D center distance as the matching criteria when counting the numbers of the false positives and false negatives. Therefore, the accuracy of the rotation angles is ignored in this evaluation metric. Although the AMOTA values do not change too much, our visualization results show that including the angular velocity in the Kalman Filter state generates better and more realistic qualitative tracking results in Figure \ref{fig:visual_car}.

The NuScenes Tracking Challenge organizer shared the test set result of the top 3 participants and the AB3DMOT \cite{weng2019ab3dmot} baseline, as in Table \ref{tab:test_results}. The full tracking challenge leaderboard will be released to public soon by the organizer.

\begin{figure*}[h!]
        \centering
        \begin{subfigure}[b]{0.3\textwidth}
            \centering
            \includegraphics[width=\textwidth]{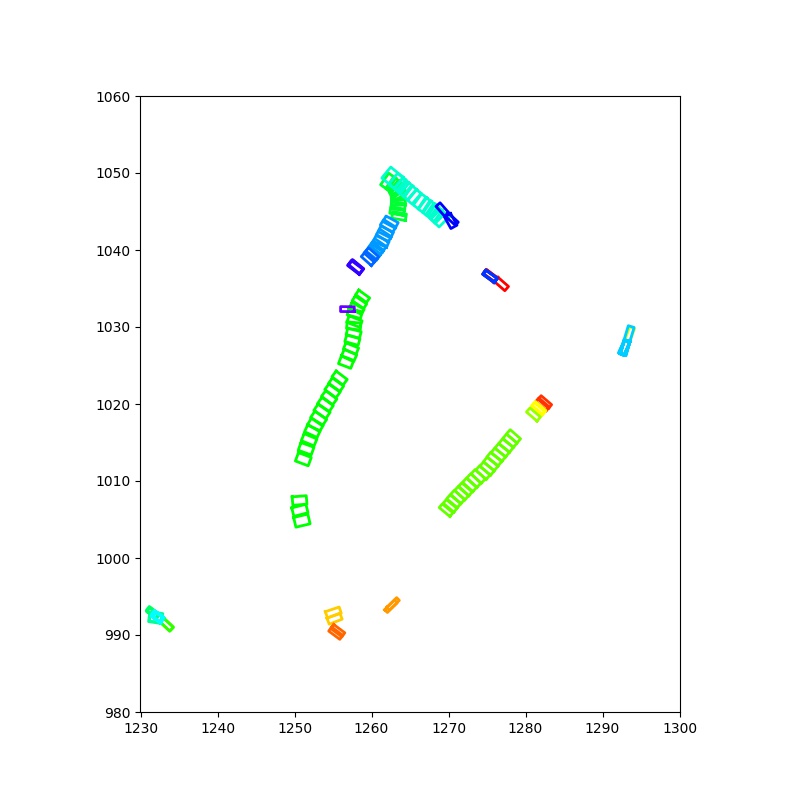}
            \caption[]%
            {{\small AB3DMOT \cite{weng2019ab3dmot}}}    
            \label{fig:pedestrian_weng2019ab3dmot_0}
        \end{subfigure}
        \hfill
        \begin{subfigure}[b]{0.3\textwidth}  
            \centering 
            \includegraphics[width=\textwidth]{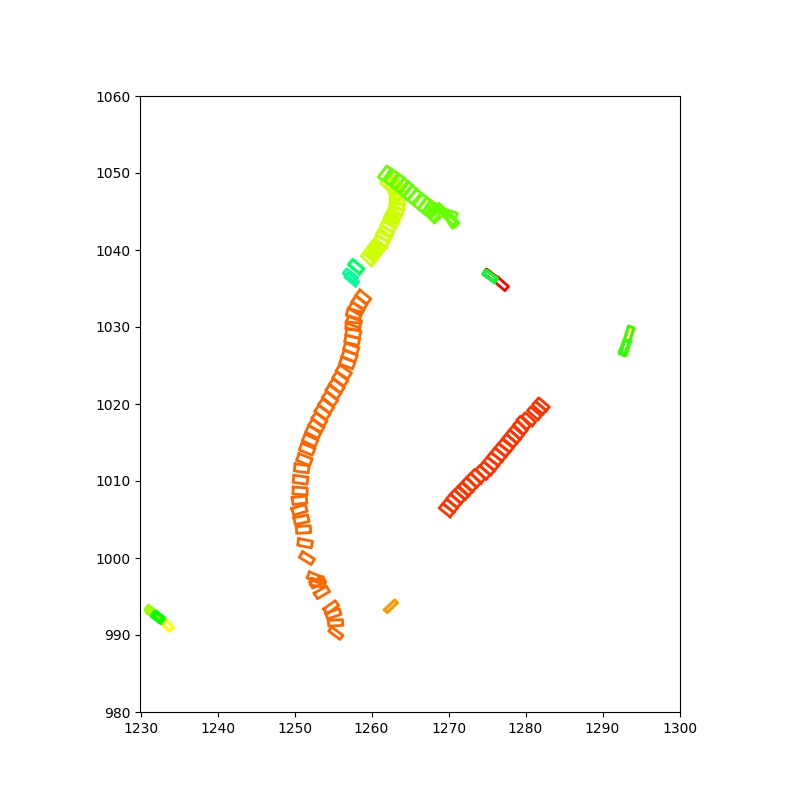}
            \caption[]%
            {{\small Ours}}    
            \label{fig:pedestrian_ours}
        \end{subfigure}
        \hfill
        \begin{subfigure}[b]{0.3\textwidth}
            \centering 
            \includegraphics[width=\textwidth]{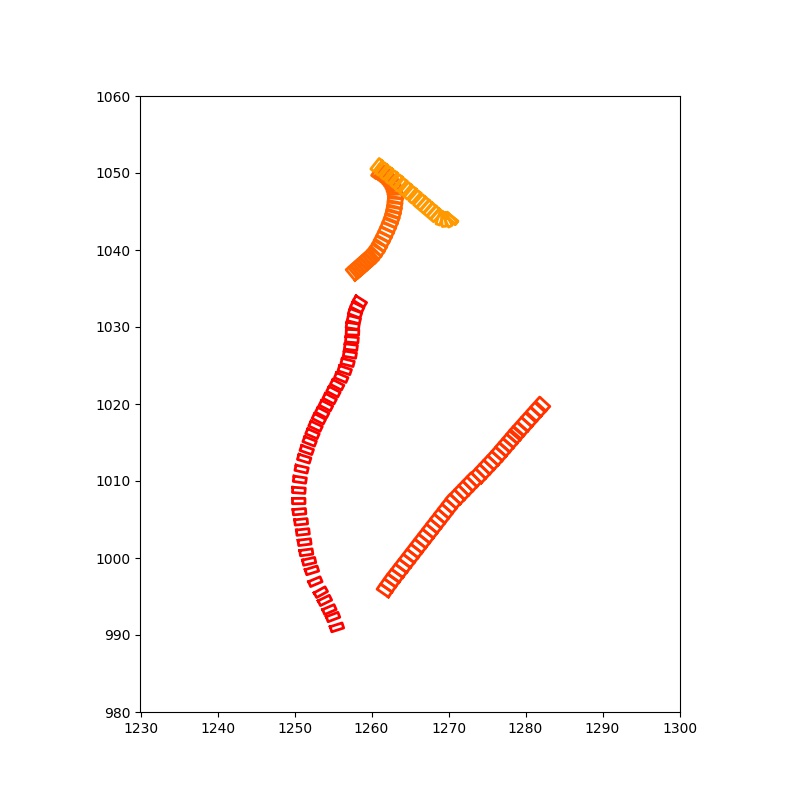}
            \caption[]%
            {{\small Ground-truth}}    
            \label{fig:pedestrian_gt}
        \end{subfigure}
        \vskip\baselineskip
        \begin{subfigure}[b]{0.3\textwidth}
            \centering 
            \includegraphics[width=\textwidth]{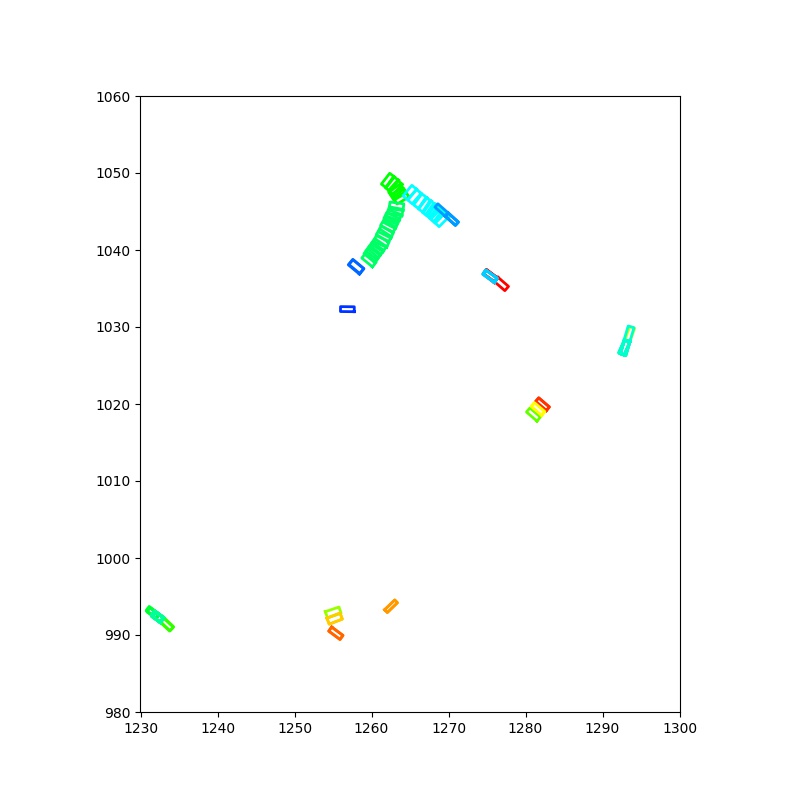}
            \caption[]%
            {{\small Our AB3DMOT \cite{weng2019ab3dmot} baseline}}    
            \label{fig:pedestrian_weng2019ab3dmot_1}
        \end{subfigure}
        \hfill
        \begin{subfigure}[b]{0.3\textwidth}
            \centering 
            \includegraphics[width=\textwidth]{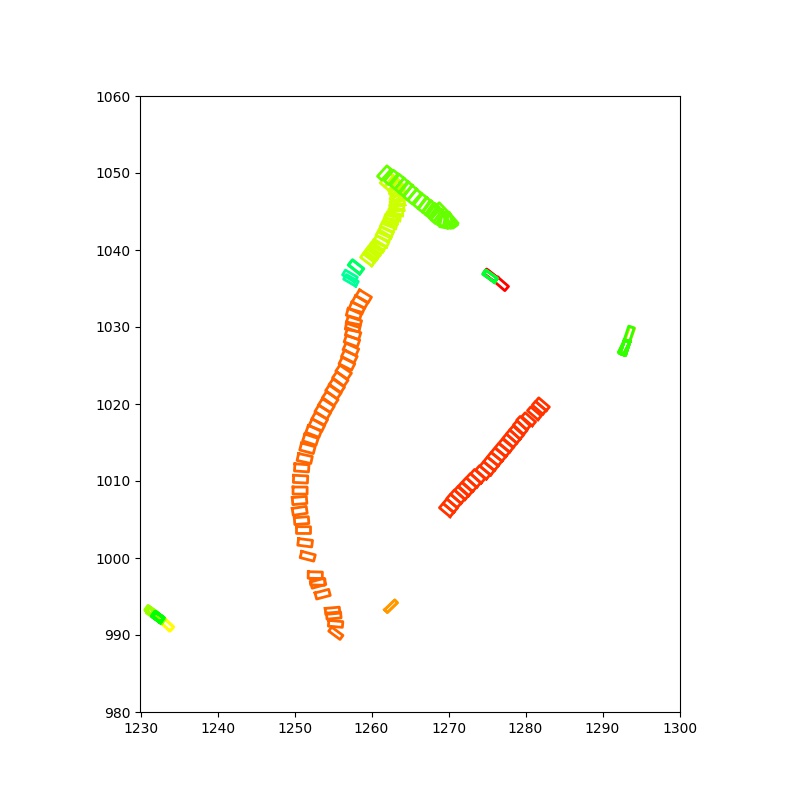}
            \caption[]%
            {{\small Ours without angular velocity}}    
            \label{fig:pedestrian_ours_without_angular_velocity}
        \end{subfigure}
        \hfill
        \begin{subfigure}[b]{0.3\textwidth}
            \centering 
            \includegraphics[width=\textwidth]{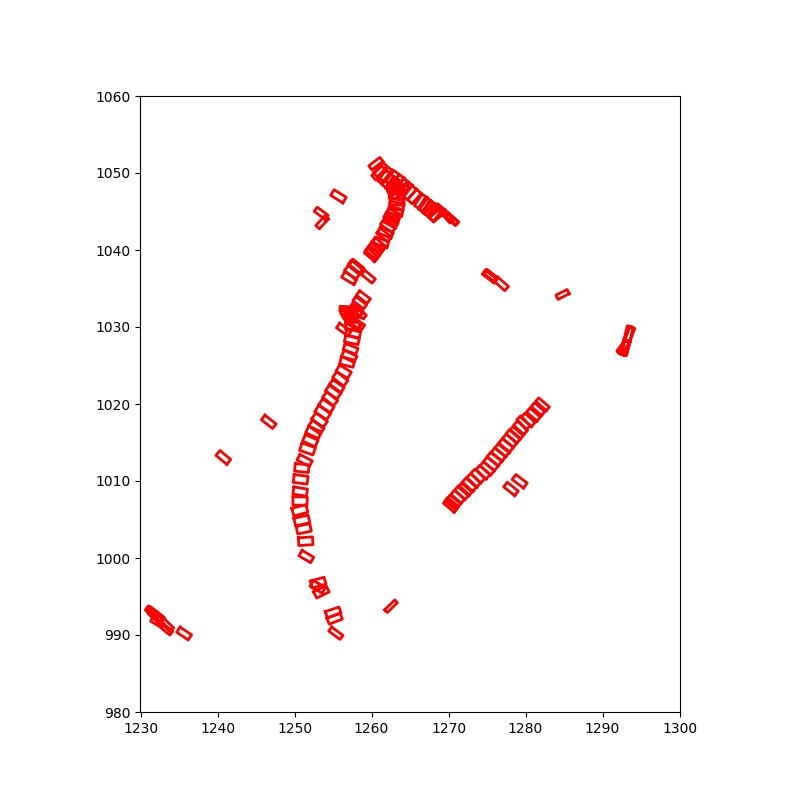}
            \caption[]%
            {{\small Input detection from MEGVII \cite{zhu2019megvii}}}    
            \label{fig:pedestrian_detection}
        \end{subfigure}
        \caption[]
        {\small Bird-eye-view tracking visualization of pedestrians} 
        \label{fig:visual_pedestrian}
        
    \end{figure*}

\subsection{Qualitative Results}
We show the AB3DMOT \cite{weng2019ab3dmot} baselines and our method's bird-eye-view visualization tracking results of the car category in Figure \ref{fig:visual_car}. We also show the ground-truth annotation and the input detection from MEGVII \cite{zhu2019megvii} as the reference. We draw the object bounding boxes from different timesteps of the same scene in a single plot. Different colors represent different instances of tracks or objects. The detection results only have a single color because no tracking id information is available.

We can see that the AB3DMOT \cite{weng2019ab3dmot} has difficulties to continue tracking when the object makes a sharp turn, as shown in Figure \ref{fig:car_weng2019ab3dmot_0} and \ref{fig:car_weng2019ab3dmot_1}. This is because the Kalman Filter's predicted 3D bounding box does not overlap with any detection box when the car is turning sharply. However, our Mahalanobis distance-based methods can still correctly track the car's motion as shown in Figure \ref{fig:car_ours} and \ref{fig:car_ours_without_angular_velocity}, either with or without angular velocity in the Kalman Filter state. For the case without using the angular velocity, the estimated car orientation during turning is obviously different from the detection or the ground-truth, as shown in Figure \ref{fig:car_ours_without_angular_velocity}, \ref{fig:car_detection}, and \ref{fig:car_gt}. Such an issue can be fixed by including the angular velocity in the Kalman Filter state as in our final proposed model, as shown in Figure \ref{fig:car_ours}. 

We show the bird-eye-view visualization for pedestrians in Figure \ref{fig:visual_pedestrian}. In this example, we can see that the input detection \ref{fig:pedestrian_detection} has some noise in the lower end of the longest track, potentially due to occlusions. The AB3DMOT \cite{weng2019ab3dmot} baseline as visualized in Figure~\ref{fig:pedestrian_weng2019ab3dmot_0} is unable to continue tracking the pedestrian. However, both of our proposed methods either with or without angular velocity (Figure \ref{fig:pedestrian_ours} and  \ref{fig:pedestrian_ours_without_angular_velocity}) can correctly track the pedestrian's location and orientation.

\section{Conclusion}
We present an on-line 3D multi-object tracking method using the Mahalanobis distance in the data association step. Moreover, we use the statistics from the training set to estimate and initialize the Kalman Filter's covariance matrices. Our method better utilizes the stochastic information and outperforms the 3D-IOU-based AB3DMOT \cite{weng2019ab3dmot} baseline by a large margin in terms of the AMOTA evaluation metric in the NuScenes Tracking Challenge \cite{caesar2019nuscenes}.   

\section{Acknowledgement}
Toyota Research Institute ("TRI") provided funds to assist the authors with their research but this article solely reflects the opinions and conclusions of its authors and not TRI or any other Toyota entity.

{\small
\bibliographystyle{ieee_fullname}
\bibliography{egbib}
}

\end{document}